# Flight Time Prediction for Fuel Loading Decisions with a Deep Learning Approach

Xinting Zhu[1], Lishuai Li[2*]

*Abstract:*

Under increasing economic and environmental pressure, airlines are constantly seeking new technologies and optimizing flight operations to reduce fuel consumption. However, the current practice on fuel loading, which has a significant impact on aircraft weight and fuel consumption, has yet to be thoroughly addressed by existing studies. Excess fuel is loaded by dispatchers and (or) pilots to handle fuel consumption uncertainties, primarily caused by flight time uncertainties, which cannot be predicted by current Flight Planning Systems (FPS). In this paper, we develop a novel spatial weighted recurrent neural network model to provide better flight time predictions by capturing air traffic information at a national scale based on multiple data sources, including Automatic Dependent Surveillance - Broadcast (ADS-B), Meteorological Aerodrome Reports (METAR), and airline records. In this model, a spatial weighted layer is designed to extract spatial dependences among network delay states (i.e. average flight delay at each airport and average flight delay of each Origin-Destination (OD) pair for a specific time interval). Then, a new training procedure associated with the spatial weighted layer is introduced to extract OD-specific spatial weights and then integrate into one model for a nationwide air traffic network. Long short-term memory (LSTM) networks are used after the spatial weighted layer to extract the temporal behavior patterns of network delay states. Finally, features from delays, weather, and flight schedules are fed into a fully connected neural network to predict the flight time of a particular flight. The proposed model was evaluated using one year of historical data from an airline's real operations. Results show that our model can provide a more accurate flight time predictions than baseline methods, especially for flights with extreme delays. We also show that, with the improved flight time prediction, fuel loading can be optimized and resulting reduced fuel consumption by 0.016% - 1.915% without increasing the fuel depletion risk.

## 1 Introduction

Many studies have been conducted to improve fuel efficiency of commercial airline flights, and most have focused on aircraft or engine designs, flight operations (e.g., optimizing altitude and speed, changing taxi procedures), and maintenance procedures. However, only a few studies have investigated unnecessary fuel loadings in flight planning (Kang & Hansen, 2018; Ryerson, Hansen,

---

[1] PhD student, Department of Systems Engineering and Engineering Management, City University of Hong Kong, Hong Kong SAR. Email: xtzhu3-c@my.cityu.edu.hk

[2] Assistant Professor, Department of Systems Engineering and Engineering Management, City University of Hong Kong, Hong Kong SAR. Email: lishuai.li@cityu.edu.hk

* Corresponding Author

& Bonn, 2014; Ryerson, Hansen, Hao, & Seelhorst, 2015). In the current practice of airline operations, the unused fuel remaining in the tank after landing is significant, which results in a large amount of waste from carrying this "dead weight" on the aircraft. Ryerson et al. found that 4.48% of the fuel consumed by an average flight is due to carrying unused fuel, and the annual cost of this unused fuel is about $230 million for the entire fleet of a major U.S. airline (Kang, Hansen, & Ryerson, 2018). Fig. 1 shows the fuel loading of a flight from Hong Kong to an airport in mainland China with the same scheduled departure time in 2017 that used the current fuel loading practice.

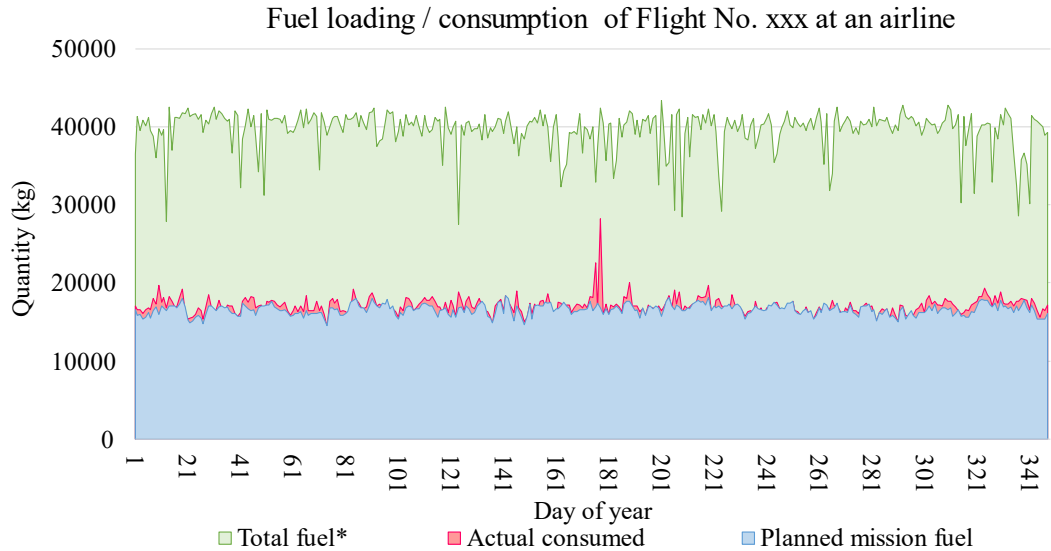

*Total Fuel Loading may include tankering fuel which is loaded due to economic considerations (e.g., it is more expensive to buy the tankering fuel at the destination to complete the return route than carrying it from the origin airport.).

Fig. 1. Fuel loading and consumption of a flight for an airline in 2017.

In the current fuel planning process, the total fuel loading of a flight consists of various fuel categories, e.g. mission fuel, alternate fuel, reserve fuel, contingency fuel, etc. In this study, in order to show how we can reduce unnecessary fuel loading by better predicting flight time, we group the various fuel categories into three categories: **Mission Fuel**, **Fixed Fuel**, and **Variable Fuel**, as shown in Fig. 2.

**Mission Fuel** represents the fuel needed to complete a planned route and is calculated by Flight Planning Systems (FPS), while Fixed Fuel and Variable Fuel describe the fuel added to handle the uncertainties of flight abnormality, such as diversion or holding, and ensure safety.

**Fixed Fuel** is determined by aviation regulations set by International Civil Aviation Organization (ICAO), Federal Aviation Administration (FAA), or other regulators, including Alternate Fuel and Reserve Fuel. Alternate Fuel refers to the fuel needed to fly from the destination airport to an alternate airport. Reserve Fuel is the amount of fuel needed to hold in the air for 45 min at normal cruising speed by Federal Aviation Regulations (FAR), or for 30 min at 450 m (1,500 ft) above the alternate aerodrome by ICAO rules.

**Variable Fuel** is defined as the sum of Contingency fuel, Extra fuel, and Discretionary fuel used in current airlines' practice. The definition of Variable Fuel in this study is same as the term 'Discretionary fuel' used in (Kang & Hansen, 2018). Contingency Fuel is required by regulations, but the quantity is normally determined by prescriptive method (a fixed percentage) or statistical method based on historical data following airline own policy. The method adopted by some airlines is called Statistical Contingency Fuel (SCF) (Kang & Hansen, 2018). Extra Fuel and Discretionary Fuel are determined by dispatchers or pilots based on their expectation of a particular flight's specific conditions, which include weather information, air traffic conditions, and other economic or engineering considerations.

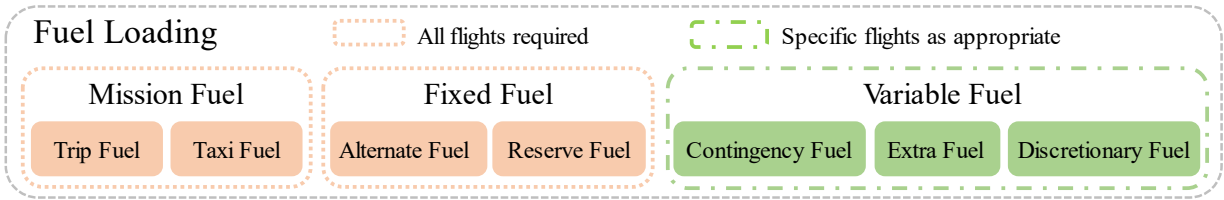

Fig. 2.  Basic fuel loading categories for current airline flight planning.

Loading Variable Fuel is a way to handle the limited capability of current FPS procedures to predict actual flight time, reflecting the dispatchers' or pilots' re-adjustment of FPS-calculated quantities of fuel loads. As shown in Fig. 1, most flights carry much more fuel than necessary because of a few records of high fuel consumption associated with significant flight time delays in the past. Thus, better predictions for flight times, especially those with significant delays, could help to reduce the need for Variable Fuel and aid in the development of better fuel loading strategies.

In this paper, in order to reduce the amount of Variable Fuel loaded on each flight, we focus on the prediction of actual flight time for individual flights, especially the flights with extreme delays. With Automatic Dependent Surveillance – Broadcast (ADS-B) being adopted by an increasing number of airlines worldwide and aviation meteorological services becoming more accessible, it is possible to collect historical and current aircraft movement data and airport weather information on a global scale. We propose a novel flight time prediction model based on deep learning approaches to extract network-based information from historical data sources and predict flight time for individual flights using real-time data. Our model is structured to be able to process inputs from multiple data sources of national air traffic systems and extract the spatiotemporal correlations among features. In addition, data availability from the perspective of airlines is considered in the model. Furthermore, a two-step training procedure is introduced to extract OD-specific features, perform model integrating, and obtain a final model that is more robust and has lower variance.

This paper is organized as follows. Section 2 reviews the existing research on flight time prediction and deep learning approaches adopted in the field of traffic prediction. Section 3 presents the proposed deep learning model for flight time prediction of individual flights. Section 4 discusses the experimental results, and a detailed analysis on potential benefits on fuel consumption is presented in Section 5. Finally, Section 6 summarizes the concluding remarks and limitations of this paper.

## 2 Literature Review

Given the nascent status of the field, research directly addressing excessive fuel loading in flight fuel planning is very rare. Ryerson et al., based on data from major U.S. airlines, estimated that 0.70% -1.04% of aircraft fuel consumption is attributed to carrying unnecessary contingency and alternate fuel (Ryerson et al., 2015). Kang et al. proposed to use flight gate-in fuel to measure aviation system predictability and estimated the cost to carry such gate-in fuel for six major U.S. airlines in 2012, which is $59 million to $667 million (Kang et al., 2018). Following this, they have also proposed a method based on quantile regression to recommend Contingency Fuel amount in the fuel planning stage (Kang & Hansen, 2017). Although there are many factors contributing to fuel consumption uncertainties, flight time has a strong correlation with the fuel consumption. Thus, this paper focuses on predicting flight time accurately.

The flight time prediction problem, as well as a related topic: flight delay prediction, have been studied for years (Mueller & Chatterji, 2002; Sternberg, Soares, Carvalho, & Ogasawara, 2017). Many prior works employ statistical methods or probabilistic approaches. Pathomsiri et al. assessed the efficiency of US airports using a non-parametric function to model joint production of on-time and delay performance (Pathomsiri, Haghani, Dresner, & Windle, 2008). Tu et al. studied major factors that influenced flight departure delays and estimated departure delay distributions at Denver International Airport for United Airlines using a probabilistic approach (Tu, Ball, & Jank, 2008). These works attempted to separate and analyze factors contributing to flight delays, but they mainly focused on one single airport, and not considered network effects. In other works, operational research, including optimization, simulations, and queue theory, has been carried out to model and simulate flight delays and help policy makers to optimize at a system level. Pyrgiotis et al. (Pyrgiotis, Malone, & Odoni, 2013) studied the delay propagation within a large network of major U.S. airports by building an analytical queuing and network decomposition model. In addition, Arıkan et al. developed stochastic models to analyze the propagation of delays through air traffic networks using empirical data (Arıkan, Deshpande, & Sohoni, 2013). Furthermore, considering the complex distributed and element-interacted properties, commercial aviation systems have been studied according to network representation. For example, Xu et al. proposed a Bayesian network approach to estimate delay propagation at three airports in the United States (Xu, Donohue, Laskey, & Chen, 2005).

These classic methods are valuable for understanding root causes and interactions among the elements of delay occurrence. However, the accuracy of these models is not sufficient for the individual flight predictions (Yu, Guo, Asian, Wang, & Chen, 2019). With the vast volume of commercial aviation system data being collected and the development of artificial intelligence algorithms, machine learning has become popular in the field of flight time prediction. The commonly used data-driven methods include k-Nearest-Neighbors, neural networks, support vector machine, fuzzy logic, and tree-based methods (Sternberg et al., 2017). Rebollo et al. adopted random forest algorithms to predict departure delays with air traffic network characteristics as input features, which had an error of approximately 21 min when predicting departure delays for a two-hour forecast horizon (Rebollo & Balakrishnan, 2014). Choi et al. combined flight schedules and weather forecasts to predict whether a scheduled flight will be delayed or on-time using several machine learning algorithms (Choi, Kim, Briceno, & Mavris, 2016). Perez-Rodriguez et al. presented an asymmetric logit probability model to estimate and predict the daily probabilities of delays in aircraft arrivals (Pérez–Rodríguez, Pérez–Sánchez, & Gómez–Déniz, 2017).

More recent, studies have used deep learning algorithms to improve the accuracy of flight delay prediction. Kim et al. proposed a recurrent neural network (RNN) to predict the flight departure and arrival delays of an individual airport with day-to-day sequences (Kim, Choi, Briceno, & Mavris, 2016). The results of this study show that the accuracy of RNN improved with deeper architectures. However, this model can only perform two-class predictions of flight delay rather than time prediction. Yu et al. analyzed high-dimensional data from Beijing International Airport and employed a novel deep belief network (DBN) with the support vector regression (SVR) method to predict flight delay at that airport. This model achieved a high accuracy with a mean absolute error (MAE) as low as 8.41 min, though the model can only be trained and used for flight delay prediction at a specific airport (Yu et al., 2019).

A common limitation of current flight delay prediction methods is that they focus on the overall performance, placing more emphasis on normal flights while neglecting the outliers (flights with significant delays), either intentionally or unintentionally. For example, Yu et al. eliminated outliers with extreme values of delays for the top and bottom 1% (Yu et al., 2019). Tu et al. excluded the "extreme" observations in data preparation to reduce the influence on smoothing spline approach (Tu et al., 2008). In (Chen & Li, 2019; Rebollo & Balakrishnan, 2014), the authors adopted random forest methods for its low sensitivity to outliers. This "optimizing for the average" technique does not serve the purpose of the problem addressed in this paper – accurately predicting flight times of individual flights in a national network, especially flights with significant delays, to reduce the need for Variable Fuel in fuel planning stage.

Recent developments of state-of-the-art model architectures of RNN units, Long-short-term-memory (LSTM) is promising to address our problem. It has been applied in the field of traffic prediction because of its ability to learn the temporal correlations of time series features. Ma et al. used LSTM for traffic speed prediction by using historical speed data from traffic microwave detectors and compared different approaches with a LSTM neural network in terms of accuracy and stability (Ma, Tao, Wang, Yu, & Wang, 2015). Liu et al. proposed a deep generative model that consists of LSTM layers as the encoder and decoder to predict aircraft trajectories (Liu & Hansen, 2018). LSTM has also been demonstrated to be effective in learning temporal correlations of time series features for many other applications (e.g., speech recognition (Graves & Jaitly, 2014), language translation (Sutskever, Vinyals, & Le, 2014), image captioning (Vinyals, Toshev, Bengio, & Erhan, 2015), video captioning (Gao, Guo, Zhang, Xu, & Shen, 2017) and signals prediction (Liang, Ke, Zhang, Yi, & Zheng, 2018).

## 3 Methodology

In this section, we present the proposed flight time prediction model, which we call the Spatial Weighted Recurrent Neural Network (SWRNN). This model can better capture extreme flight time delays by combining comprehensive aviation information from different data sources and extracting key influential factors from original features with proper structures.

The key challenge of developing such a data-driven flight time prediction model lies in how to incorporate the network level spatiotemporal features and flight specific features into a single model to predict individual flight time. The air traffic network is a complex distributed transportation system, and its interacted elements lead to flight time delays directly or inductively (Rebollo & Balakrishnan, 2014; Sternberg et al., 2017). Particular model structures need to be carefully designed to extract useful information from the spatiotemporal and high-dimensional data inputs.

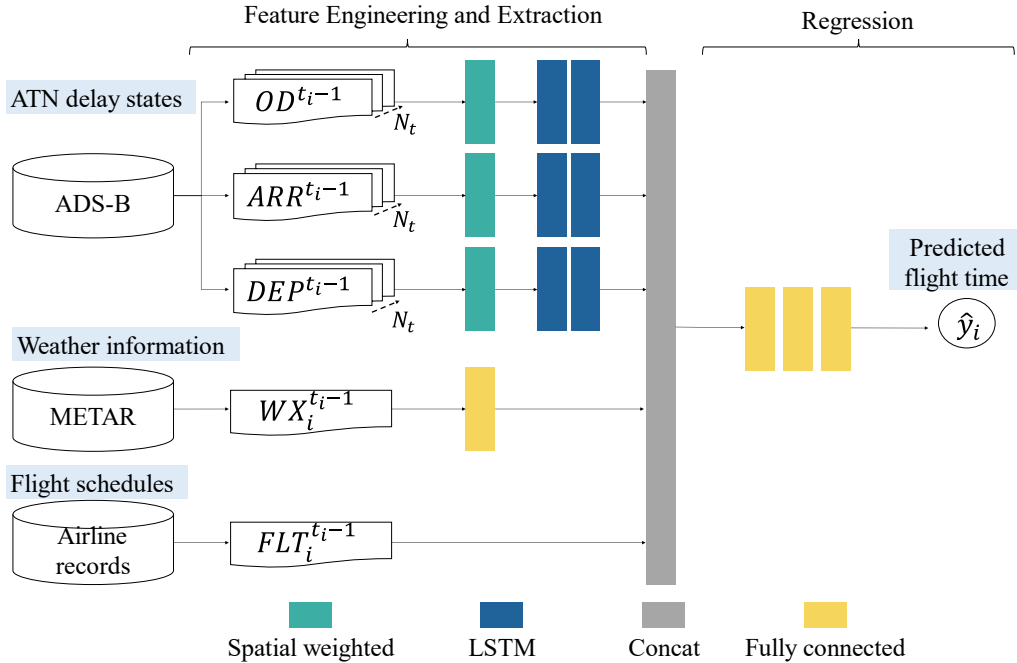

Fig. 3. Framework of SWRNN model.

As the model framework illustrated in Fig. 3, three kinds of features are fed to the different structured branches in the extraction module after engineering from raw data sources. To extract the dynamic networkwide traffic conditions, the main components of SWRNN's extraction module are a spatial weighted layer and two adopted recurrent neural network layers. The spatial weighted layer (SWL) is designed for dimension reduction, which is the process of extracting useful spatial features from initial inputs that contain information on multiple airports and origin-destination (OD) pairs. The recurrent neural network layers are structured to extract temporal correlations between features, identifying repetitive traffic patterns and interacting elements in air traffic network. Moreover, the training procedure of OD-specific SWL is introduced. After pre-training OD-specific spatial weights to extract the spatial dependency, we combine all samples together to retrain and get one integrated model for the entire air traffic network. For weather information, a fully connected layer is adopted in the extraction module.

This framework incorporates dynamic air traffic information on network delay status. Experiments in section 4 demonstrate the proposed model's superior performance on capturing extreme flight time delays. After concatenating the useful features extracted from original complex structured and high-dimensional inputs, a fully connected neural network is adopted for regression and outputs the predicted flight time.

## 3.1 Model output and inputs

The model output is the predicted enroute flight time of sample flight $i$, which is denoted as $\hat{y}_i$. Suppose the scheduled departure time for the sample flight $i$ is denoted as $t_i$. The time that the model makes a prediction is $t_i - 1$, one time step ahead the scheduled departure time. The model inputs include three types of information, air traffic network (ATN) delay states, weather information, and flight schedule information, as listed in Table 1.

The ATN delay states describe the average delay time of flights by OD-pair, arrival airport, and departure airport in the past $N_t$ time steps, which are represented by matrix $OD^{t_i-1} \in \mathbb{R}^{N_t \times N_{OD}}$, $Arr^{t_i-1} \in \mathbb{R}^{N_t \times N_{AP}}$, and $Dep^{t_i-1} \in \mathbb{R}^{N_t \times N_{AP}}$, respectively. $N_{OD}$ is the total number of OD pairs and $N_{AP}$ is the total number of airports in the studied ATN. The ATN delay states are calculated using ABS-D data.

Weather information specifies the weather condition of the origin airport of flight $i$ at time $t_i - 1$ and the weather condition of the destination airport of flight $i$ at its scheduled arrival time. A vector $W_i^{t_i-1}$ with $2N_{WX}$ dimensions contains such information, where $N_{WX}$ is the dimension of a standard set of variables decoded from the Meteorological Aerodrome Report (METAR).

Flight schedule information comes from the airline records. It includes origin and destination airports, aircraft type, planned flight time, scheduled arrival and arrival times, hour-of-day, day-of-week, month, and departure and arrival demand for corresponding airports. The input features of flight information are selected from airline records, which provide basic information of each flight to be predicted.

Table 1 Summary of model inputs

To predict the flight time $\hat{y}_i$ of sample flight $\underline{i}$ at time $t_i - 1$, the model takes the following inputs:

| Category (data source) | Notation | Remarks |
|---|---|---|
| ATN delay states (ADS-B) | $OD^{t_i-1} = (x_{OD_k}^{t_i-j}) \in \mathbb{R}^{N_t \times N_{OD}}$ | OD delay states in the past $N_t$ time steps; $x_{OD_k}^{t_i-j}$ is the average flight delay on route $OD_k$ in time step $t_i - j, j = \{1,2,\dots,N_t\}, OD_k = \{OD_1, OD_2, \dots, OD_{N_{OD}}\}$ |
| | $ARR^{t_i-1} = (x_{Arr_k}^{t_i-j}) \in \mathbb{R}^{N_t \times N_{AP}}$ | Arrival airport delay states in the past $N_t$ time steps; $x_{Arr_k}^{t_i-j}$ is the average arrival delay at destination airport $Arr_k$ in time step $t_i - j, j = \{1,2,\dots,N_t\}, Arr_k = \{Arr_1, Arr_2, \dots, Arr_{N_{AP}}\}$ |
| | $DEP^{t_i-1} = (x_{Dep_k}^{t_i-j}) \in \mathbb{R}^{N_t \times N_{AP}}$ | Departure airport delay states in the past $N_t$ time steps; $x_{Dep_k}^{t_i-j}$ is the average departure delay at origin airport $Dep_k$ in time step $t_i - j, j = \{1,2,\dots,N_t\}, Dep_k = \{Dep_1, Dep_2, \dots, Dep_{N_{AP}}\}$ |

| Weather information (METAR) | $WX_i^{t_i-1} \in \mathbb{R}^{2N_{WX}}$ | Weather conditions of the origin airport of flight $i$ at time $t_i - 1$, and weather conditions of the destination airport of flight $i$ at scheduled arrival time |
|---|---|---|
| Flight schedule information (Airline records) | $FLT_i^{t_i-1} \in \mathbb{R}^{N_{FLT}}$ | Origin airport, Destination airport, Planned flight time, Aircraft type, Scheduled departure time, Scheduled arrival time, Hour-of-day, Day of week, Month, Departure demand at origin, Arrival demand at destination |

## 3.2 Feature Engineering and Extraction

We develop specific feature engineering and extraction methods to process the three groups of input features, ATN delay states, weather information and flight schedule information, and then incorporate them together for the next regression module to predict flight time for an individual flight.

### 3.2.1 ATN delay states

Aviation system is a highly interacted network, where air traffic control (ATC) procedures are wide-implemented and influenced multi-stakeholders and elements in the network. For example, if an aircraft cannot arrive as scheduled, the reasons may come from many aspects: this aircraft departs late at origin or holds in air because of the congested conditions of destination; Sudden weather changes or sectors' blockades en-route also lead to the traffic congestions. What's more, the departure late of this aircraft may not only influenced by the conditions of origin and/or destination airports but the up-stream flights' delay. This sequential effect is called delay propagation, and this kind of 'late-arriving' takes a big share of flight delay causes in air traffic (Bureau of Transportation Statistics, 2016; Wang, Zheng, Wu, Chen, & Hansen, 2020).

This interacted network effect implies us to include the spatial-dependent features on network delay information in the prediction model. Besides, the historical temporal correlations, such as delay patterns within ATN like peak hours or weekend effects, will also be useful in the traffic prediction problems.

### a) Definition

Inspired by the definition of network delay characteristic used in (Rebollo & Balakrishnan, 2014), we adopted the feature ATN delay states to describe the historical delay information for the entire aviation network. ATN delay states include three categories, OD-pair delay states, airport arrival delay states, and airport departure delay states, each of which is a time series with $N_t$ timesteps as inputs. Therefore, for a flight $i$ to be predicted at one hour before its scheduled departure time $t_i$, the adopted features of the three ATN delay states estimated from time $t_i - N_t$ to time $t_i - 1$ are denoted as $OD^{t_i-1} = \left(x_{OD_k}^{t_i-j}\right) \in \mathbb{R}^{N_t \times N_{OD}}, ARR^{t_i-1} = \left(x_{Arr_k}^{t_i-j}\right) \in \mathbb{R}^{N_t \times N_{AP}}, DEP^{t_i-1} = \left(x_{Dep_k}^{t_i-j}\right) \in \mathbb{R}^{N_t \times N_{AP}}$, where $j = \{1, 2, \dots, N_t\}$.

The spatiotemporal network delay features are counted on an hourly basis from historical ADS-B data. For each timestep $t_i - j$, the element $x_{OD_k}^{t_i-j}$ in the OD-pair delay states is defined as the average gate delays for flights belong to OD-pair $j$ arrived at the time interval $[t_i - j - \Delta t, t_i - j)$. Here $\Delta t$ is indicated as the unit of time interval, and $\Delta t = 1$ since it is counted hourly. The value of $x_{OD}^j$ is calculated by

$$x_{OD_k}^{t_i-j} = \frac{1}{n} \sum_{f=1}^{n} \left( \left( A_{arr}^f - A_{dep}^f \right) - \left( S_{arr}^f - S_{dep}^f \right) \right),$$

$$s.t., f \in \Omega \left( A_{arr}^f \in [t_i - j - \Delta t, t_i - j) \right) \cap \Omega \left( OD_f = OD_k \right), \tag{1}$$

where $A_{arr}^f$, $A_{dep}^f$ denotes the actual arrival/departure time of flight $f$, and $S_{arr}^f$, $S_{dep}^f$ denotes the scheduled arrival/departure time of flight $k$; $k$ denotes the $k^{th}$ OD-pair in ATN.

The $x_{Arr_k}^{t_i-j}$, $x_{Dep_k}^{t_i-j}$ in airport arrival delay states and airport departure delay states are defined as the average arrival delay and average departure delay of $k^{th}$ Airport in ATN at time interval $[t_i - j - \Delta t, t_i - j)$, where $\Delta t = 1$. Given that $ARR^k, DEP^k$ indicates the arrival and departure airports of flight $k$, the calculation for $x_{Arr_k}^{t_i-j}, x_{Dep_k}^{t_i-j}$ are given by

$$x_{Arr_k}^{t_i-j} = \frac{1}{n} \sum_{f=1}^{n} \left( \left( A_{arr}^f - S_{arr}^f \right) \right),$$

$$s.t., f \in \Omega \left( A_{arr}^f \in [t_i - j - \Delta t, t_i - j) \right) \cap \Omega \left( ARR_f = ARR_k \right) \tag{2}$$

$$x_{Dep_k}^{t_i-j} = \frac{1}{n} \sum_{F=1}^{n} \left( \left( A_{dep}^f - S_{dep}^f \right) \right),$$

$$s.t., f \in \Omega \left( A_{dep}^f \in [t_i - j - \Delta t, t_i - j) \right) \cap \Omega \left( DEP_f = DEP_k \right). \tag{3}$$

**b) Extraction module**

To extract useful features and reduce feature dimensions of ATN delay states, we adopt (i) a spatial weighted layer to learn OD-pair/airports spatial dependencies from the high-dimension input sequences, (ii) Stacked LSTM to extract temporal correlation of time series of OD-pair/airports delay states.

*i) Spatial weighted layer (SWL)*

This layer is designed to extract the OD-specific feature and address the high dimensionality problem for the feature ATN delay states. The learnable weights of the spatial weighted layer show the importance of different OD-pair/airports to the sample flight. For example, the part of OD-pair delay states is given by

$$\widehat{OD}^{t_i-1} = \sigma \left( \mathbf{1}^{N_t} * \left( \boldsymbol{W}_{OD}^l \right)^\top \otimes OD^{t_i-1} + \boldsymbol{b}_{OD}^l \right), \tag{4}$$

where $\widehat{OD}^{t_i-1} = \left(\hat{x}_{OD_k}^{t_i-j}\right) \in \mathbb{R}^{N_t \times N_{OD}}$. In the vector $\mathbf{1}^{N_t} \in \mathbb{R}^{N_t}$, each element equals to one. In addition, $\boldsymbol{W}_{OD}^l \in \mathbb{R}^{N_{OD}}$, $\boldsymbol{b}_{OD}^l \in \mathbb{R}^{N_{OD}}$ are learnable variables that are learned in the first step of training procedures, where $l \in \{1, 2, \ldots, N_{OD}\}$ is chosen by the value of OD-pair of sample flight $i$. Finally, $\otimes$ indicates the Hadamard product and $*$ indicates matric multiplication. Besides, the $\sigma$ denotes the activation function, where we adopted the LeakyReLu function. The spatial weight $w_{OD,k}^l \in \mathbb{R}$ indicates the $k^{th}$ element in $\boldsymbol{W}_{OD}^l$, and its value represents the importance of each contributing feature of OD-pair delay states.

The parts for airport arrival and departure delay states are similar and denoted as

$$\widehat{ARR}^{t_i-1} = \sigma \left( \mathbf{1}^{N_t} * \left(\boldsymbol{W}_{ARR}^l\right)^\top \otimes ARR^{t_i-1} + \boldsymbol{b}_{ARR}^l \right) \tag{5}$$

$$\widehat{DEP}^{t_i-1} = \sigma \left( \mathbf{1}^{N_t} * \left(\boldsymbol{W}_{DEP}^l\right)^\top \otimes DEP^{t_i-1} + \boldsymbol{b}_{DEP}^l \right) \tag{6}$$

where $\boldsymbol{W}_{ARR}^l, \boldsymbol{W}_{DEP}^l \in \mathbb{R}^{N_{AP}}, \boldsymbol{b}_{ARR}^l, \boldsymbol{b}_{DEP}^l \in \mathbb{R}^{N_{AP}}$ are learnable variables to be trained in the first step of training procedures and $w_{ARR,k}^l, w_{DEP,k}^l \in \mathbb{R}$ indicate the elements in $\boldsymbol{W}_{ARR}^l, \boldsymbol{W}_{DEP}^l$, respectively. The variable $l \in \{1, 2, \ldots, N_{OD}\}$ is determined by the OD-pair of sample $i$, which keeps the same with the part for OD-pair delay states in (1). Demonstration of spatial weighted layer is shown in Fig. 4.

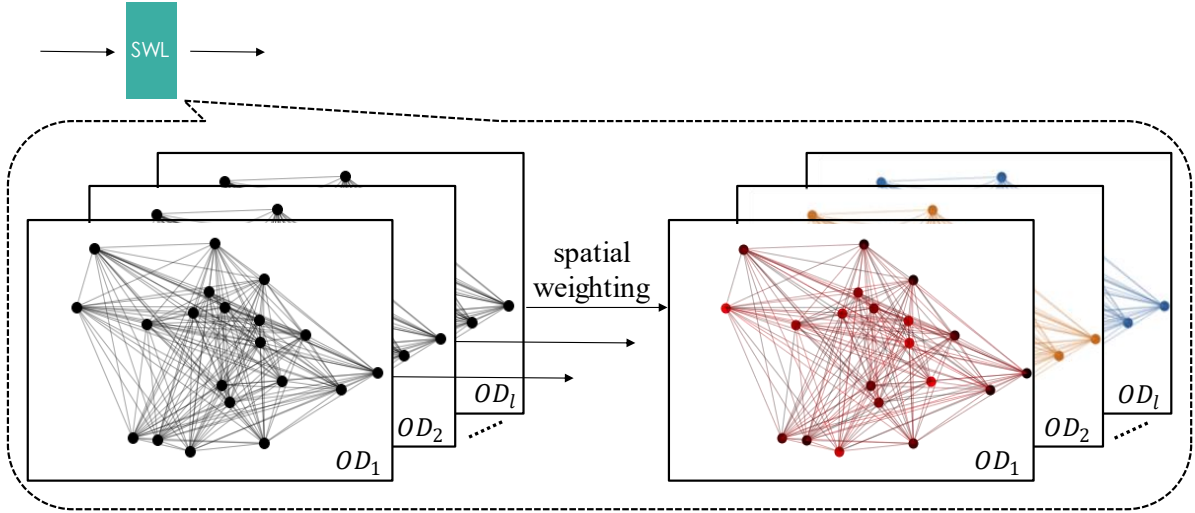

Fig. 4. Demonstration of SWL's function.

## ii) LSTM layer

We choose to use LSTM layer in the model because it has been shown that LSTM is effective in learning long and short temporal dependences (Zhao, Chen, Wu, Chen, & Liu, 2017). Denoting the cell state as $c_t \in \mathbb{R}^M$ and the hidden state as $h_t \in \mathbb{R}^M$, where $M$ is the dimension of the LSTM layer, the mechanism can be executed with the steps as demonstrated in Fig. 5. Moreover, we use stacked LSTM (the number of layers is denoted as $q$) as the unit of feature extraction blocks to enhance model performance. Here we set $q$ as 2, which is a common setting. The learning ability of one layer is not enough, while three layers are too expensive and complex to learn with limited contributions to the overall performance. The stacked LSTM generates a set of outputs for each kind of ATN delay states. These three sets of outputs are denoted as $O_{LSTM}(\widehat{OD}^{t_i-1}), O_{LSTM}(\widehat{ARR}^{t_i-1}), O_{LSTM}(\widehat{DEP}^{t_i-1})$, respectively.

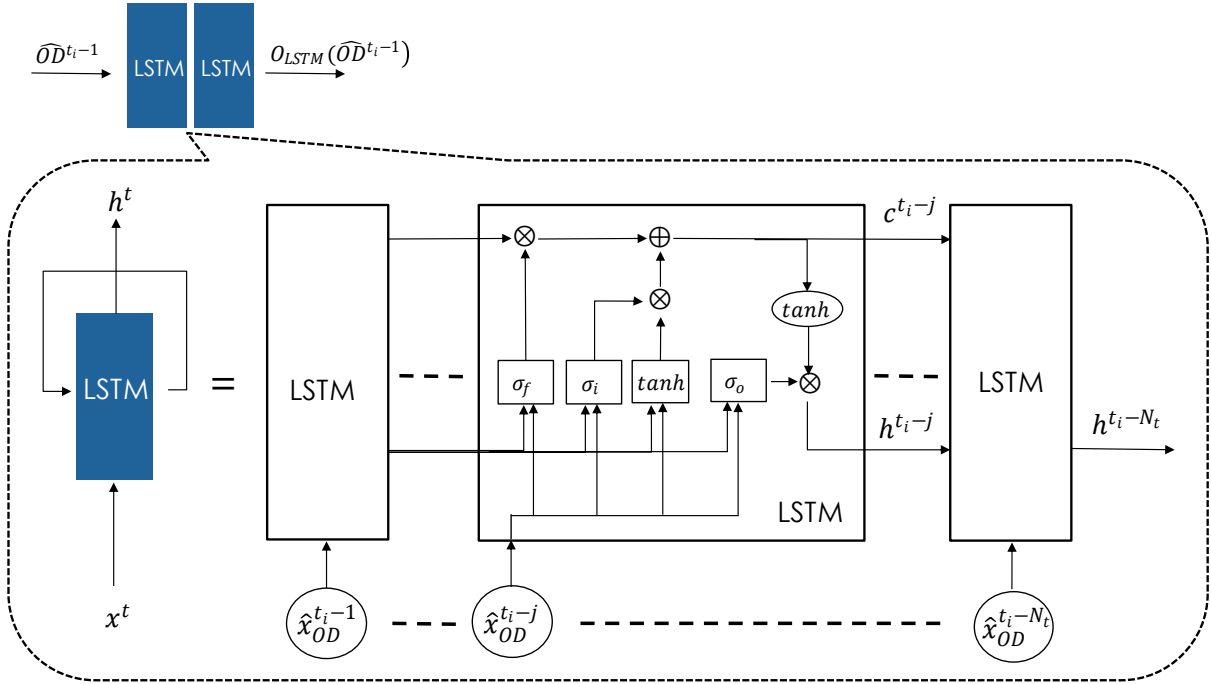

Fig. 5. Demonstration of LSTM mechanism. Take OD-pair delay states as example.

### c) Training OD-specific SWL

Since the mixture distributions existing in the spatial dependencies of ATN delay states, OD-pair is selected as the key to cluster the distributions. However, for each OD-pair, the flight samples are limited, and individual OD-specific models suffer from overfitting problems in the deep learning approach. Thus, we introduced a two-step training procedure to generate OD-specific SWL at the first step and then combine them into a final integrated SWRNN model for the entire network. An illustration of this two-step training procedure is presented in Fig. 6.

We first pretrain OD-specific SWL ($SWL\_l$) using flight samples from the same OD-pair $l$ in the training set to obtain its OD-specific spatial dependencies within ATN delay states features, i.e., the learnable parameters of three SWL $\boldsymbol{W}_{OD}^l, \boldsymbol{b}_{OD}^l, \boldsymbol{W}_{ARR}^l, \boldsymbol{b}_{ARR}^l, \boldsymbol{W}_{DEP}^l, \boldsymbol{b}_{DEP}^l$ are corresponding to specific OD-pair $l$ of flight samples as mentioned in (4-6). Then in the second step, we freeze the weights of SWL learned in the first step, merge the samples together, and shuffle to train the other parts of the SWRNN model. The pretrained technique is commonly adopted in deep network or transfer learning to extract meaningful features from a large-scale dataset. Here we borrow the idea but transfer it to train the multiple OD-specific models at first and then combine to train the other parts together to enhance the model performance. Since the limited training samples for each of the OD models, the two-step training can obtain a more robust generic model and reduce over-fitting. The superior performance on model accuracy of the two-step training with OD-specific SWL rather than the one-time united training is further analyzed in the experiment.

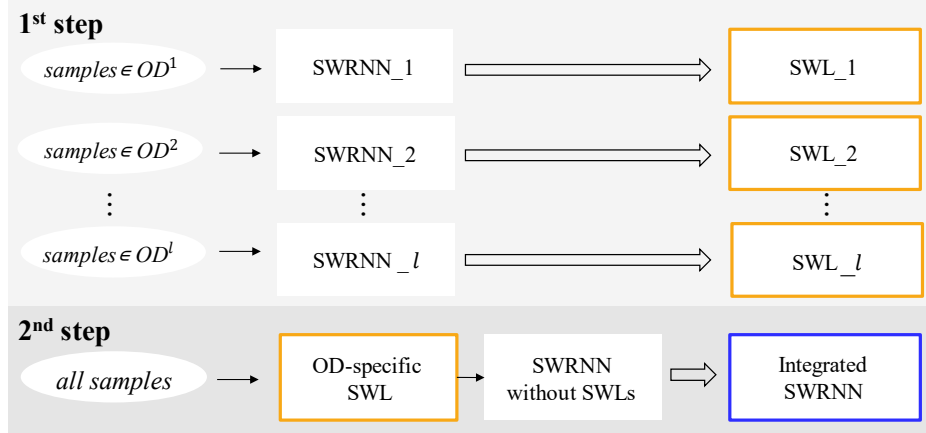

Fig. 6. Illustration of training procedures with OD-specific SWL.

### 3.2.2 Weather information

Updated weather forecasts can reflect the real-time airport capacity. In addition to extreme weather conditions, wind gust, snow, and fog etc. also influence terminal efficiency, which affects flight delays directly. METAR data is utilized as weather reports on terminals for airlines and pilots' reference in daily aviation operations.

### a) METAR decoding

METAR is a formatted text report released every half-hour or an hour, which contains the terminal weather information on wind, cloud, temperature, humidity, etc. Fig. 7 shows an example of raw METAR. After collecting the data for each airport containing in our dataset, we decoded the raw text based METAR data into numerical values by meteorology parameter (e.g., 10 knots for wind speed). Adopted weather information of SWRNN model include a standard set of meteorology variables. For a flight $i$ scheduled to depart at time $t_i$, adopted inputs on weather information is real-time, at $t_i - 1$ at departure airport of flight $i$, and the weather information at destination airport at scheduled arrival time of flight $i$.

$$WX_i^{t_i-1} = \left[ X_{Dep_i}^{t_i}, X_{Arr_i}^{t_i'} \right] \in \mathbb{R}^{2N_{WX}} \tag{7}$$

Where $t_i'$ indicates the scheduled arrival time of flight $i$. $X$ represents the adopted variables, including wind direction, wind speed, wind gust, cloud type, cloud height, visibility and whether it is in visual meteorological conditions (VMC). The feature 'Whether it is VMC' is determined according to the visual flight rules (VFR) by related values decoding from METAR.

| Raw format | METAR VHHH 010000Z 08011KT 9999 FEW022 SCT028 20/14 Q1022 NOSIG= |
|---|---|
| Decoded report | Location.........................................: VHHH<br>Day of month..............................: 01<br>Time............................................: 00:00 UTC<br>Wind...........................................: True direction = 080 degrees; Speed = 11 knots<br>Visibility ....................................: 10 km or more<br>Cloud coverage............................: Few (1 to 2 oktas) at 2200 feet above aerodrome level<br>Cloud coverage............................: Scattered (3 to 4 oktas) at 2800 feet above aerodrome level<br>Temperature.................................: 20 degrees Celsius<br>Dewpoint.....................................: 14 degrees Celsius<br>QNH (msl pressure)......................: 1022 hPa<br>Next 2 hours................................: No significant changes |

| Adopted features | Wind Direction (degree) | Wind Speed (knots) | Wind Gust (knots) | Cloud Type | Cloud Height (feet) | Visibility (meters) | VMC (Y/N) |
|---|---|---|---|---|---|---|---|
| Values | 80 | 11 | 0 | Few | 2200 | 9999 | Y |

Fig. 7. A METAR example of VHHH airport in the dataset.

We did not include the enroute weather information in this model due to a lack of data. However, the enroute condition can be partially reflected by OD-pair delay states.

**b) Extraction module**

We adopt a fully connected layer (FCL) to learn and select useful features from the comprehensive weather information, and its mechanism expressed as

$$\widehat{WX}_i^{t_i-1} = \boldsymbol{W}_{WX} * WX_i^{t_i-1} + \boldsymbol{b}_{WX}, \qquad (8)$$

where $\boldsymbol{W}_{WX} \in \mathbb{R}^{2N_{WX} \times N_h}, \boldsymbol{b}_{WX} \in \mathbb{R}^{N_h}$ are learnable during training, $N_h$ is the hidden dimension of the FCL and its output $\widehat{WX}_i^{t_i-1} \in \mathbb{R}^{N_h}$.

### 3.2.3 Flight information

The features of flight information provide the basic information that characterizes the flight to be predicted, which is selected based on domain knowledge. The specific features of flight information are listed in Table 1, including Origin airport, Destination airport, Aircraft type, Scheduled departure time, Scheduled arrival time, Planned flight time, Hour-of-day, Day of week, Month, Departure demand at origin, Arrival demand at destination. They are engineered from airline records and ADS-B data. The features Arrival demand and Departure demand, which are collected from the ADS-B data, indicate the number of flights for which the scheduled arrival and departure times from the corresponding airports are within one-hour intervals before the prediction time.

### 3.3 Regression

After feature concatenating, the output is

$$\widehat{X}_{t_i-1}^i = \left[ O_{LSTM}\left(\overline{OD}^{t_i-1}\right), O_{LSTM}\left(\overline{ARR}^{t_i-1}\right), O_{LSTM}\left(\overline{DEP}^{t_i-1}\right), \widehat{WX}_i^{t_i-1}, FLT_i^{t_i-1} \right]. \qquad (9)$$

Next, a fully connected neural network, also known as a multilayer perceptron (MLP), is utilized in the regression module. Here we adopt a 3-layer MLP. The output predicted time of sample flight $i$ is then

$$\hat{y}_i = MLP\big(\widehat{\boldsymbol{X}}_{t_{i-1}}^i\big). \tag{10}$$

Furthermore, the techniques of batch normalization (BN) and dropout are adopted. BN is used to increase the stability of a neural network, which normalizes the output of a previous activation layer by subtracting the batch mean and dividing by the batch standard deviation (Ioffe & Szegedy, 2015). Dropout is a regularization technique that randomly selects ignored neurons during training (Nitish Srivastava, Geoffrey Hinton, Ilya Sutskever, 2014). The techniques are commonly used to reduce overfitting in deep learning.

Since our approach is smooth and differentiable, SWRNN can be trained via the backpropagation algorithm (David E. Rumelhart, Geoffrey E. Hintont, 1986). During the training phase, we use the Adam optimization algorithm (Kingma & Ba, 2015) to train our model by minimizing the mean squared error (MSE) between the predicted $\hat{y}_i$ and the ground truth $y_i \in \mathbb{R}$ via

$$Loss(\theta) = \|\hat{y}_i - y_i\|^2, \tag{11}$$

where $\theta$ denotes all the learnable parameters in the proposed model.

## 4 Evaluation and Testing

### 4.1 Data Description

This study requires three different sources of data: airline records, ADS-B data, and METAR data.

We obtained a set of airline records from a Hong Kong based airline. The records contain information on scheduled and actual departure and arrival times, fuel loadings and actual fuel consumption of each flight. This dataset includes passenger flights operated between Hong Kong and mainland China cities from January 1st to December 31st 2017, involving 41 OD-pair and 20 airports.

The ADS-B data contain scheduled and actual times of all flights operating in the ATN during the year 2017. These flights include not only the OD pairs in the ATN, but also other international OD pairs so that the ATN delay states can be counted more accurately. We specify the scope of the ATN in the experiment as a simplified Chinese aviation network, which consists of 53 airports and 549 OD pairs (i.e., $N_{OD} = 549, N_{AP} = 53$). This simplified ATN is selected from OD pairs that have more than six flights per day on average, which approximately represents real traffic conditions in Chinese aviation at the national scale.

The METAR data are collected according to origin and destination airports involved in the flight samples, i.e. the 20 airports in the airline records during the year 2017.

Missing data are deleted, accounting for 0.02% of all data. Categorical features are encoded by one-hot encoding. Then the data are normalized by scaling each column to have a mean of zero and standard deviation of one. After this preprocessing, the number of flights is 39067 and the total dimension of model inputs is 15814.

## 4.2 Evaluation and Testing Setup

### 4.2.1 Training-validation-test split

Given that we focus on the prediction accuracy for outlier flights with extreme flight time delays in this study, we first separate the flights into an outlier set and a normal set, so that we can evaluate the model performance on outliers and normal flights separately. Then we partition the data into non-overlapped training, validation and test data by a ratio of 3:1:1. The outlier set and the normal set are sampled equally: 60% from both the outlier and normal sets and 60% to form the training set, 20% from both the outlier and normal sets to form the validation set, and the remaining 20% from both the outlier and test sets to form the test set.

The outlier set is defined as

$$i \in \begin{cases} outlier\ set, & \left|FDT_i - \mu_g\right| > 2\sigma_g \\ normal\ set, & \left|FDT_i - \mu_g\right| \leq 2\sigma_g \end{cases}, \tag{12}$$

where $FDT_i$ denotes the flight delay time of flight $i$, and $\mu_g$ and $\sigma_g$ are the mean and standard deviation, respectively, of the delay times of flight group $g$ having the same OD-pair and aircraft type as flight $i$. Here the coefficient of $\sigma_g$ is determined by the trade-off between physical meaning and the number of flights. In our dataset, the number of selected outlier flights is 1330 with 21 minutes average flight delay and 18 minutes delay standard deviation. Table 2 summarizes the number of flight samples in each set.

Table 2
Summary of sample sets

| Number of samples | All | Normal | Outlier |
|---|---|---|---|
| Training set | 23471 | 22671 | 800 |
| Validation set | 7766 | 7507 | 259 |
| Test set | 7830 | 7559 | 271 |

### 4.2.3 Benchmarks

To compare with SWRNN, we tested three additional methods: FPS, least absolute shrinkage and selection operator (LASSO) and recurrent neural network (RNN). FPS, which is the adopted model in current airline practice, generates the planned flight time by historical statistic and physical calculations (this also works as an input of flight information features for other Benchmarks). LASSO is a commonly used method for problems with high dimensional feature inputs. Originally, it is formulated for least squares models and used the L1 penalty as regularization in order to enhance the prediction accuracy. RNN denoted here has the same model structure with SWRNN except the spatial weighted layers, which is an ablation study of SWRNN to check the structure effectiveness of the designed SWL. The inputs for LASSO and RNN are the same as the ones used for SWRNN.

#### 4.2.4 Hyperparameters

During the training procedure, we set the batch size as 256 and the learning rate as 0.001, which is a common setting for Adam. The hidden dimensionality of each layer is decided by grid search, the final values of which are shown in Table 3.

Table 3
Hyperparameter settings

| Layer name | Number of units |
|---|---|
| SWL for OD-pair delay states | 549 |
| SWL for airport arrival delay states | 53 |
| SWL for airport departure delay states | 53 |
| Stacked LSTM for OD-pair delay states | 40 |
| Stacked LSTM for airport arrival delay states | 10 |
| Stacked LSTM for airport departure delay states | 10 |
| FCL for weather information | 10 |
| FCL1 for MLP | 150 |
| FCL2 for MLP | 100 |
| FCL3 for MLP | 30 |
| Output layer | 1 |

The number of timesteps ($N_t$) is set via sensitivity analysis. We test RMSE performance when $N_t \in \{2, 6, 12, 24, 36, 48\}$, as shown in Fig. 8. A value of $N_t$=24 is chosen as the number of timesteps in the following evaluation.

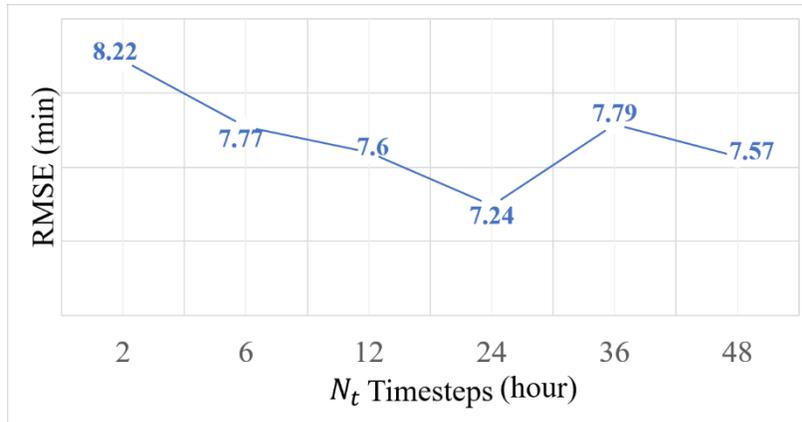

Fig. 8. Timestep sensitivity analysis for SWRNN in validation set.

### 4.3 Evaluation Result

To evaluate the model performance of the flight time prediction, we adopt three standard metrics including Root Mean Square Error (RMSE), MAE, and R-squared ($R^2$). Table 4 summarizes the performance of the four methods on the outlier set and the normal set. SWRNN (refers to with OD-specific SWL training if no other notes) consistently demonstrated better performance than FPS, LASSO and RNN, especially in the outlier set. Compared to FPS, SWRNN improved flight time predictions by about 9 min according to the RMSE and by about 10 min according to the MAE. Compared to LASSO, the SWRNN results improved the predictions by about 5 min according to the RMSE and by 6 min according to the MAE. For normal flights, SWRNN showed limited improvement in prediction accuracy. This is because the current planned flight times predicted by FPS are sufficiently accurate and have limited potential for improvement. LASSO performed slightly better than FPS and much worse than SWRNN. LASSO is well known for its ability to manage high-dimensional data. However, it has limited ability to utilize complex structured feature inputs. RNN, as an ablation study, shows worse performance than SWRNN, which indicates the effectiveness of spatial weighted layer to extract the spatial correlations within ATN delay states features.

Besides, the comparison of SWRNN on OD-specific SWL training shows that the proposed two-step with OD-specific SWL training outperformed the regular one-time united training, for both normal and outlier flights. With the OD-specific SWL training, the model performance is enhanced by better capturing the OD-specific features within nationwide ATN delay states. Due to the limited training samples for each OD pair, it is not realistic to build an OD-specific model for each OD pair. In contrast, a general model for the whole network without OD-specific SWL training provides results with reduced accuracy. Therefore, the two-step training procedure provides a good balance between prediction accuracy and model complexity.

Table 4
Performance on three kinds of flights in test set

(a) Outlier

| | FPS | LASSO | RNN | SWRNN (without OD-specific SWL training) | SWRNN |
|---|---|---|---|---|---|
| RMSE (min) | 26.56 | 22.52 | 18.31 | 18.48 | 17.82 |
| MAE (min) | 24.20 | 20.29 | 14.89 | 15.23 | 14.50 |
| $R^2$ | 0.6136 | 0.7223 | 0.8163 | 0.8131 | 0.8260 |

(b) Normal

| | FPS | LASSO | RNN | SWRNN (without OD-specific SWL training) | SWRNN |
|---|---|---|---|---|---|
| RMSE (min) | 8.47 | 6.96 | 7.73 | 7.80 | 6.90 |
| MAE (min) | 6.50 | 5.49 | 5.22 | 5.19 | 5.31 |
| $R^2$ | 0.9544 | 0.9692 | 0.9651 | 0.9614 | 0.9698 |

(c) All

| | FPS | LASSO | RNN | SWRNN (without OD-specific SWL training) | SWRNN |
|---|---|---|---|---|---|
| RMSE (min) | 9.68 | 8.02 | 8.32 | 8.40 | 7.55 |
| MAE (min) | 7.11 | 6.00 | 5.56 | 5.54 | 5.63 |
| $R^2$ | 0.9411 | 0.9596 | 0.9565 | 0.9557 | 0.9642 |

Fig. 9 shows the distributions of predicted flight time errors for FPS and SWRNN. For FPS, the mean and standard deviation of the predicted errors are -3.14 and 9.16, respectively. For SWRNN, they are -0.41 and 7.54, respectively. Fig. 10 shows the predicted error as a function of actual flight time for FPS and SWRNN. The plot in Fig. 10(a) indicates that SWRNN provides better flight time prediction than FPS for flights with actual flight times more than 100 minutes. Additionally, for the outlier flights shown in Fig. 10(b), SWRNN's predictions have lower errors.

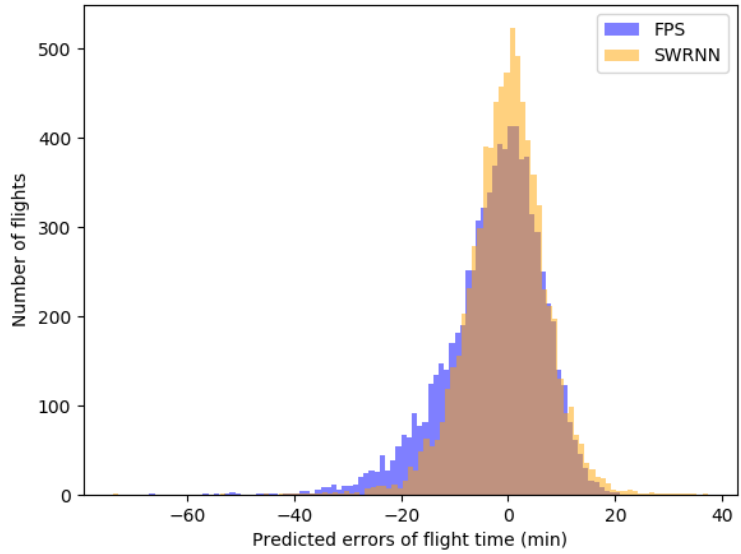

Fig. 9. Distribution of predicted errors of flight times in test set (errors are defined as the difference between predicted flight times of FPS/SWRNN and actual flight times).

(a) All

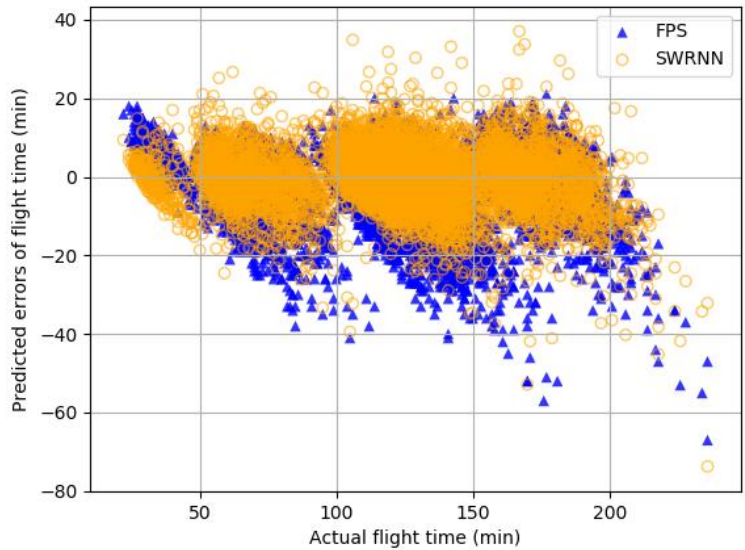

(b) Outliers

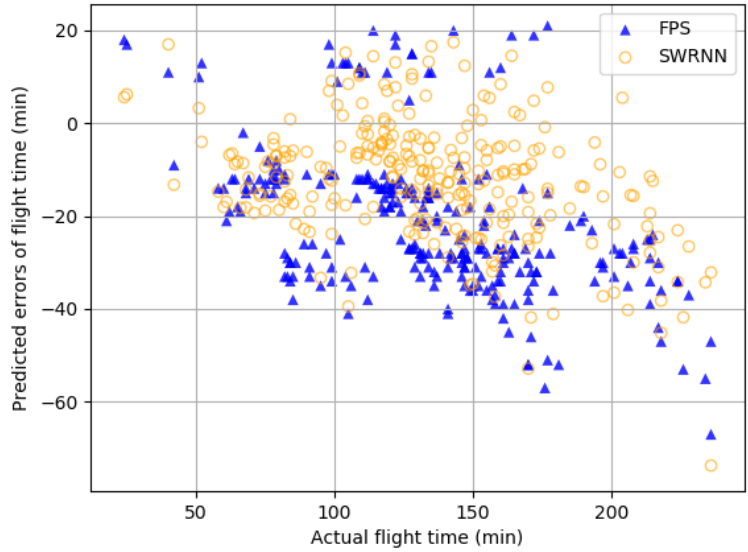

Fig. 10. Predicted errors of flight times in test set as a function of actual flight time.

## 4.4 Case Study

To further explore the model performance of SWRNN, we select a set of flights with the same origin-destination-departure hour (i.e., the same flight number). We compare the actual flight time with the FPS planned flight time and the SWRNN predicted flight time. The results are shown in Fig. 11, which shows that the actual flight time peaks can be more accurately captured by SWRNN than FPS in general.

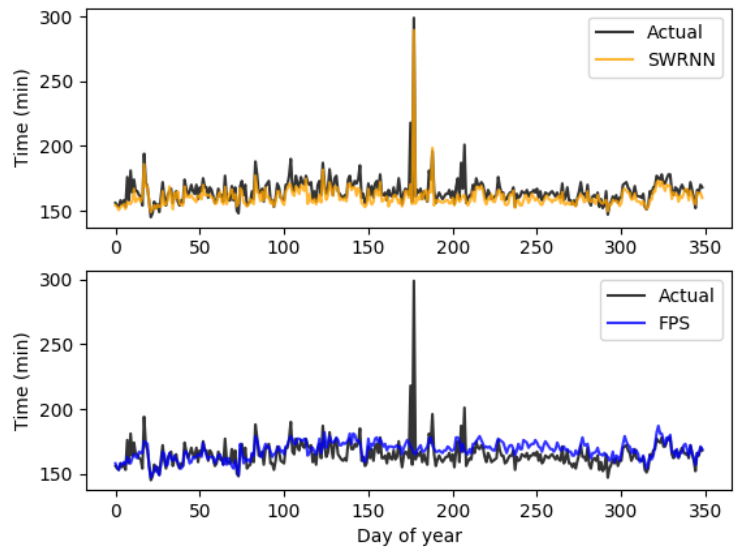

Fig. 11. Comparison of actual flight times and predictions by SWRNN (top) and FPS (bottom) for an example flight number on each day of an entire year.

Fig. 12 is plotted to exam if there is any relationship between actual flight time and the prediction error. We observe that with longer actual flight time, FPS tends to underestimate the flight time more severely, because current FPS only provides additional fuel for exceptional weather/traffic situation but flight time, however is not adjusted to reflect the effect due to weather/traffic, while SWRNN is able to correct that trend slightly. In Fig.9, f3 in the plot is an example that SWRNN performed much better than FPS, while f1 and f2 are examples of both SWRNN and FPS failed to capture the actual flight time. This may suggest that these two cases could be caused by factors not considered in the modelling. Reason for these extreme cases needs further exploration. A summary of the model performance on the example flight is shown in Table 5.

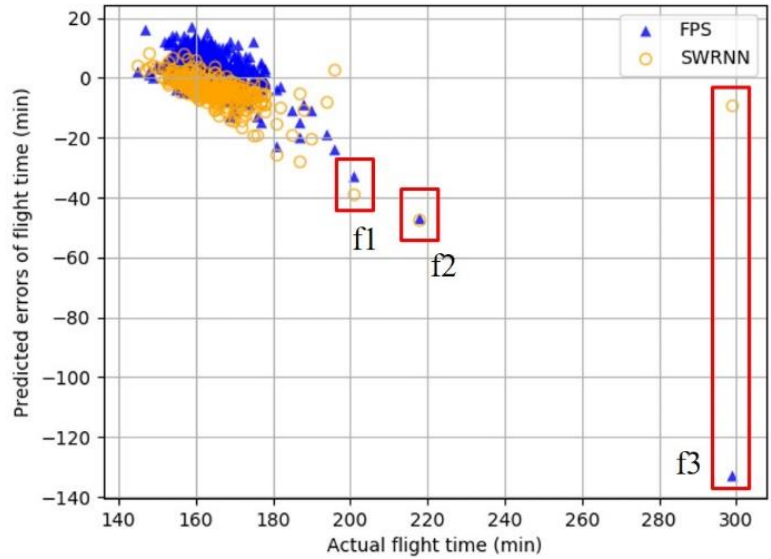

Fig. 12. Predicted errors of flight times for the specific flight number as a function

Table 5
Performance of the specific flight in case study

|  | FPS | SWRNN | SWRNN |
|---|---|---|---|
|  |  | (without OD-specific SWL training) |  |
| RMSE (min) | 10.55 | 10.34 | 6.92 |
| MAE (min) | 6.42 | 5.81 | 5.04 |
| $R^2$ | 0.0679 | 0.1243 | 0.5992 |

# 5 Fuel consumption analysis

In this section, we explore the potential benefits to fuel consumption and the risks of fuel depletion if SWRNN is adopted to make fuel loading decisions. As a proof-of-concept analysis, we propose a hypothetical fuel loading strategy based on SWRNN output. In this hypothetical strategy, we assume dispatchers or pilots trust the proposed fuel loading based on SWRNN results. All the fuel consumption in this section excluded the influence of the tankering fuel that is loaded due to economic considerations.

## 5.1 Hypothetical fuel loading strategy

In this proof-of-concept study, we propose the fuel loading of each flight should only include:

- **Mission Fuel**, including Trip fuel and Taxi fuel, where Trip fuel is calculated based on SWRNN flight time prediction, and Taxi fuel keeps the same as it is in current practice.
- **Fixed Fuel**, which includes Alternate Fuel and Reserve Fuel as required by regulations, as it is in current practice.
- **Trip fuel buffer**, loaded for the uncertainties not captured by model, which is calculated by adding a constant buffer time $b$, and transferring it to the corresponding fuel consumption. The buffer time is hypothetically assigned as 10 min for a pro-efficiency policy, and 25 min for a pro-safety policy. The b-value for the pro-safety policy is set as a rough upper bound of the fuel loading, which can maintain the same safety level as current airline practice, while the b-value for the pro-efficiency policy is set as an aggressive lower bound to assess the potential benefits without causing unacceptable risks.

Thus, the proposed fuel loading strategy can be formulated as

$$\begin{aligned}
&Fuel\ Loading \\
&= Mission\ Fuel + Fixed\ Fuel + Trip\ Fuel\ buffer \\
&= Trip\ Fuel + Taxi\ Fuel + Fixed\ Fuel + b*(Trip\ Fuel/Flight\ Time)
\end{aligned}$$

$$\text{where } b = \begin{cases} 10\ min, & if\ pro\text{-}efficiency \\ 25\ min, & if\ pro\text{-}safety \end{cases}. \tag{13}$$

## 5.2 Fuel model

The Mission Fuel in (10) is calculated based on the SWRNN predicted flight time following the procedure used in current FPS: use a factor to convert the flight time into fuel consumption. This factor will be determined by aircraft departure weight, aircraft type, and engine conditions, etc. Here we estimate the factor from our dataset directly. For each flight $i$, the factor is denoted as $\beta_i$ and its mission fuel is calculated via

$$Mission\ Fuel_i = \beta_i * (Fuel\ Loading_i + ZFW_i) * FT_i, \tag{14}$$

where $ZFW_i$ denotes the zero fuel weight of flight $i$. $FT_i$ denotes the planned flight time by FPS and predicted flight time by SWRNN, respectively. $Fuel\ Loading_i$ indicate planned total fuel loading in FPS and proposed fuel loading in (10), and $Mission\ Fuel_i$ indicate the planned mission fuel in FPS and calculated mission fuel by SWRNN, respectively.

## 5.3 Benefit and risk calculation

For benefit analysis of both policies, we evaluate how much less fuel would be carried at departure and how much less fuel would be consumed, assuming each policy was applied to the studied flights. The calculation of how much less fuel would be carried is straightforward: take the difference between the original fuel loading and the new fuel loading with the pro-efficiency policy or the pro-safety policy. How much less fuel would be consumed is calculated based on a Cost-to-Carry (CTC) factor. The CTC factor (in kg per kg·m) for a given flight is defined as the gate-to-gate fuel consumption (in kg) with its departure weight (in kg) and path distance (in m) (Ryerson et al., 2015), which can be estimated from the data directly. $Dist_i$ represents the path distance of flight $i$ and $ZFW_i$ denotes its zero fuel weight. The calculation of benefits is given by

$$Less\ carried\ fuel_i = Fuel\ Loading_i^{FPS} - Fuel\ Loading_i^{SWRNN} \qquad (15)$$

$$Less\ consumed\ fuel_i = CTC_i * Less\ carried\ fuel_i * Dist_i$$

$$where\ CTC_i = \frac{Mission\ Fuel_i^{FPS}}{Dist_i * \left(ZFW_i + Fuel\ Loading_i^{FPS}\right)}. \qquad (16)$$

The safety risk in this part of the analysis refers to the probability of Reserve Fuel being used at landing for flight $i$, defined as

$$Risk_i = \textbf{Prob}\{(Fuel\ Loading_i - Consumed\ Fuel_i) < Reserve\ Fuel_i\}, \qquad (17)$$

This equation is consistent with airline's current practice of reporting safety incidents when landing with reserve fuel being used, which should be in a very low probability rate.

### 5.3 Benefit and risk analysis results

Flights in the airline records were divided into two groups for this part of analysis: Group 1 contains all flights from Hong Kong to a city in mainland China, and Group 2 refers to the flights with the opposite direction, flying from a city in mainland China to Hong Kong, due to their distinctive fuel loading and delay characteristics. Table 6 summarizes the benefit estimation and risk assessment of the two groups of flights at the studied airline for each of the two hypothetical fuel loading policy. It is noted that all the benefit estimations in Table 6 excluded the influence of the tankering fuel that is loaded due to economic considerations.

For flights from Hong Kong to mainland China (16830 flights in 2017), with the pro-efficiency policy, the fleet would carry 55.2 million kg (15.032%) less fuel, leading to 4.3 million kg (2.844%) less fuel burned. However, safety performance would be sacrificed: 30 flights (0.178%) would be reported as incidents. With the pro-safety policy, 1 flight (0.006%) would use Reserve Fuel at landing, and 35.0 million kg (9.490%) less fuel would be carried, which would result in a reduction in fuel consumption by 2.9 million kg (1.915%).

For flights from mainland China to Hong Kong (16847 flights in 2017), the improvement potential is limited. With the pro-efficiency policy, the fleet would carry 22.0 million kg (8.737%) less fuel, leading to 1.8 million kg (1.112%) less fuel burned, and 18 flights (0.107%) would be reported as incidents. With the pro-safety policy, no flight would use Reserve Fuel at landing, and 2.2 million kg (0.878%) less fuel would be carried, which would result in a reduction in fuel consumption by 0.03 million kg (0.016%).

Table 6
Benefit estimation and risk assessment
(A) flights from Hong Kong to mainland China

| Fuel loading policy | Less carried fuel (%) | Less consumed fuel (%) | Risk (%) |
|---|---|---|---|
| Current | - | - | $0.000^1, 2.335^2$ |
| Pro-efficiency | 15.032 | 2.844 | 0.178 |
| Pro-safety | 9.490 | 1.915 | 0.006 |

| (B) flights from mainland China to Hong Kong | | | |
|---|---|---|---|
| Fuel loading policy | Less carried fuel (%) | Less consumed fuel (%) | Risk (%) |
| Current | - | - | $0.000^1, 0.000^2$ |
| Pro-efficiency | 8.737 | 1.112 | 0.107 |
| Pro-safety | 0.878 | 0.016 | 0.000 |

1 Real operations

2 After excluding the tankering fuel

The difference on benefit improvement between the two groups of flights is expected. Flights from Hong Kong to mainland China are subject to larger uncertainties in flight time (11% flights with delay larger than 15 minutes), compared to flights flying back to Hong Kong where the delay is more predictable and it is the airline's home base (3% flights with delay larger than 15 minutes). The current tools used at airlines have limited capability in predicting uncertainty in the complex and dynamic air traffic conditions. Dispatchers or pilots tend to load more fuel to make flights safe. That is why the SWRNN based fuel loading polices could save more on the outbound flights from Hong Kong than the inbound ones.

Regarding to the environmental effects and monetary savings, we utilize the U.S. Environmental Protection Agency conversion factor (EPA Center for Corporate Climate Leadership, 2018) to translate fuel savings into reduction in CO2 emission in kg, which is 9.75 kg/gallon for Kerosene-Type Jet Fuel. We assume 0.3223 kg/gallon for the fuel density and $1.6/gallon for the jet fuel price (IATA - fuel price monitor, 2019.). Our study shows that, for the entire fleet (33677 flights in 2017), if applying the pro-efficiency policy, the airline could save 1.967% on fuel consumption, translating to $30.3 million in fuel costs and 184.6 million kg of CO2 emissions. If applying the pro-safety policy, the airline could save 1.025% on fuel consumption, which is $15.8 million in fuel costs and 96.1 million kg of CO2 emissions. A summary is shown in Table 7.

Table 7

Monetary savings and reduced gas emission for entire fleet in 2017

| Fuel loading policy | Less consumed fuel (million kg, %) | Monetary savings (million $) | Reduced CO2 (million kg) |
|---|---|---|---|
| Pro-efficiency | 6.102, 1.967 | 30.290 | 184.579 |
| Pro-safety | 3.178, 1.025 | 15.778 | 96.147 |

## 6 Conclusion

In this paper, we proposed a novel deep learning flight time prediction model, SWRNN, for the purpose of reducing excessive fuel loading for scheduled airline flights. SWRNN takes the historical ATN delay states at the network scale and corresponding weather and flight information into consideration. It can extract useful features from complex structured inputs such as the non-linear spatial and temporal correlations from original data sources. An OD-specific SWL training procedure is introduced to extract OD-specific spatial dependencies and combine to obtain the final integrated SWRNN model with limited training samples. We compared the model performance with three benchmark methods, FPS, LASSO and RNN. Furthermore, we analyzed and visualized the efficacy of SWL with OD-specific training procedure. The SWRNN model shows higher accuracy with RMSE of approximately 7 min on average. Particularly for outlier flights, the prediction performance of SWRNN has shown its superior accuracy than the other methods.

We also analyzed the potential fuel savings and associated risks if SWRNN is used to inform fuel loading decisions. We estimated that 0.016% for flights from mainland China to Hong Kong and 1.915% for flights from Hong Kong to mainland China (1.025% on average for the entire fleet) in fuel savings could be realized without sacrificing safety risks in the studied airline.

These results are promising, as it shows the potential in using deep learning techniques to improve flight time predictions and inform fuel loading policies. However, there are several limitations of this study. First, SWRNN could not capture all significant delays and prediction errors exist for normal flights as well. Our future work will be more focused on improving model performance for flights with significant flight time delays while keeping the error range on normal flights within an acceptable level so that a fuel loading policy with less buffer can be developed. Second, the hypothetical fuel loading policies developed in Section V are preliminary. More work is required to optimize flight fuel loading towards a data-driven fuel planning strategy, supporting airlines to plan flights with realistic uncertainties.

## Acknowledgments

The authors would like to thank Steve Yip and his team at Cathay Pacific Airways for providing data, domain knowledge, and insightful discussions throughout this study. The work was supported by the Hong Kong Research Grants Council General Research Fund (Project No. 11215119 and 11209717).